\documentclass[sigconf]{acmart}

\usepackage{booktabs}
\usepackage{bm}
\usepackage{url}
\usepackage{mathrsfs}
\usepackage{multirow}
\usepackage{balance}
\usepackage{amsthm}
 
\usepackage{amsmath}
\usepackage{amstext}
\usepackage{amssymb}
\usepackage[ruled,vlined,linesnumbered]{algorithm2e}
\usepackage{subcaption}
\usepackage{overpic}
\usepackage{color}
\usepackage{enumitem}
\usepackage{fancyvrb}
\usepackage{graphicx}
\usepackage{caption}
\usepackage{multirow}

\DeclareMathOperator*{\argmin}{arg\,min}

\SetKwInput{KwInput}{Input}                
\SetKwInput{KwOutput}{Output}              

\renewcommand\footnotetextcopyrightpermission[1]{} 
\AtBeginDocument{%
  \providecommand\BibTeX{{%
    \normalfont B\kern-0.5em{\scshape i\kern-0.25em b}\kern-0.8em\TeX}}}

\setcopyright{acmcopyright}
\copyrightyear{2020}
\acmYear{2020}
\acmDOI{10.1145/1122445.1122456}

\begin{document}

\title{Generative Counterfactuals for Neural Networks via Attribute-Informed Perturbation}

\author{Fan Yang, Ninghao Liu, Mengnan Du, Xia Hu}
\affiliation{
  \institution{Department of Computer Science and Engineering, Texas A\&M University} 
  \city{College Station, TX, USA, 77843} 
  }
 \email{{nacoyang, nhliu43, dumengnan, xiahu}@tamu.edu} 

\renewcommand{\shortauthors}{F. Yang et al.}

\begin{abstract}
With the wide use of deep neural networks (DNN), model interpretability has become a critical concern, since explainable decisions are preferred in high-stake scenarios. Current interpretation techniques mainly focus on the feature attribution perspective, which are limited in indicating \emph{why} and \emph{how} particular explanations are related to the prediction. To this end, an intriguing class of explanations, named \emph{counterfactuals}, has been developed to further explore the ``what-if'' circumstances for interpretation, and enables the reasoning capability on black-box models. However, generating counterfactuals for raw data instances (i.e., text and image) is still in the early stage due to its challenges on high data dimensionality and unsemantic raw features. In this paper, we design a framework to generate counterfactuals specifically for raw data instances with the proposed \textbf{\underline{A}}ttribute-\textbf{\underline{I}}nformed \textbf{\underline{P}}erturbation (\textbf{AIP}). By utilizing generative models conditioned with different attributes, counterfactuals with desired labels can be obtained effectively and efficiently. Instead of directly modifying instances in the data space, we iteratively optimize the constructed attribute-informed latent space, where features are more robust and semantic. 
Experimental results on real-world texts and images demonstrate the effectiveness, sample quality as well as efficiency of our designed framework, and show the superiority over other alternatives. Besides, we also introduce some practical applications based on our framework, indicating its potential beyond the model interpretability aspect. 
\end{abstract}

\maketitle

\section{Introduction}

The past decade has witnessed the success of deep neural networks (DNN) in a wide range of application domains~\cite{pouyanfar2018survey}. Despite the superior performance, DNN models have been increasingly criticized due to its black-box nature~\cite{doshi2017towards}. Interpretable machine learning techniques~\cite{du20techniques} are thus becoming significantly vital, especially in those high-stake scenarios, such as medical diagnosis. To effectively interpret black-box DNNs, most approaches investigate the feature attributions between input instances and output predictions through correlation analysis, so that humans can have a sense of which part of the instance contributes most to the model decision. A typical example is the heatmaps employed for image classification~\cite{selvaraju2017grad}, where related saliency scores are capable of indicating the feature importance for one particular prediction label. 

However, existing correlation-based explanations are neither discriminative nor counterfactual~\cite{pearl2009causal}, since they are not able to help understand why and how particular explanations are relevant to model decisions. Thus, to further explore the decision boundaries of black-box DNN, \emph{counterfactuals} have gradually come to the attention of researchers, as an emerging technique for model interpretability. Counterfactuals are essentially some synthetic samples within data distribution, which can flip the model prediction. With counterfactuals, humans can understand how input changes affect the model and conduct reasoning under the ``what-if'' circumstances. Take a loan applicant who got rejection for instance. Correlation-based explanations may simply indicate those most contributed features (e.g., income and credit) for rejection, while counterfactuals are capable of showing how the application could be accepted with certain changes (e.g., increase the monthly income from $\$5,000$ to $\$7,000$). 

Recent work have already made some initial attempts on conducting such counterfactual analysis. The first line of research~\cite{kim2016examples,chen2019looks} employed the prototype and criticism samples in the training set as the raw ingredients for counterfactual analysis, even though those selected samples are not counterfactuals in nature. Prototypes indicate the set of data samples that best represent the original prediction label, while criticisms are the samples with the desired prediction label which are close to the decision boundary. Some other work~\cite{goyal2019counterfactual,agarwal2019removing} further utilized feature replacement techniques to create hypothetical instances as counterfactuals, where a query instance and a distractor instance are typically needed for counterfactual generation. The key of this kind of methodologies lies in the effective feature extraction and efficient replacement algorithm. Besides, contrastive intervention~\cite{dhurandhar2018explanations,white2019measurable} on the query instance is another common way to generate counterfactuals regarding to the desired label. By reasonably perturbing input features, counterfactuals can be obtained in the form of modified data samples. 

Despite the existing efforts, generating valid counterfactuals for raw data instances is still challenging due to the following reasons. First, effective counterfactuals for certain label are not guaranteed to be existed in training set, so the selected prototypes and criticisms are not always sufficient for counterfactual analysis. The related sample selection algorithms are highly possible to select some ``unexpected'' instances due to data constraints~\cite{kim2016examples}, which would largely limit the reasoning on model behaviors. Second, efficient feature replacement for raw data instances could be very hard and time-consuming~\cite{goyal2019counterfactual}. Also, relevant distractor instances for replacement may not be available in particular scenarios considering privacy and security issues, such as loan applications. Third, modifying query samples with intervention can simply work on a limited types of data, such as tabular data~\cite{white2019measurable} and naive image data~\cite{dhurandhar2018explanations}. For general raw data like real-world texts or images, intervention operation in data space can be extremely complicated and intractable, which makes it difficult to be used in practice. 

To handle the aforementioned challenges of counterfactual generation for raw instances, the high-dimension data space and unsemantic raw features are the two obstacles ahead. To this end, in this paper, we design a framework to generate counterfactuals specifically for raw data instances with the proposed \textbf{\underline{A}}ttribute-\textbf{\underline{I}}nformed \textbf{\underline{P}}erturbation (\textbf{AIP}) method. By utilizing the power of generative models, we can obtain useful hypothetical instances within the data distribution for counterfactual analysis. Essentially, our proposed AIP method can guide a well-trained generative model to generate valid counterfactuals by updating its parameters in the attribute-informed latent space, which is a joint embedding space for both raw features and data attributes. Compared with the original input space, attribute-informed latent space has two significant merits for counterfactual generation: (1) raw features are embedded as low-dimension ones which are more robust and efficient for generation; (2) data attributes are modeled as joint latent features which are more semantic for conditional generation. As for the construction of attribute-informed latent space, we typically employ two types of losses to conduct the training of generative models, where the reconstruction loss is used to guarantee the quality of raw feature embedding and the discrimination loss is used to ensure the correct attribute embedding. Through the gradient-based optimization, the proposed AIP method can iteratively derive the valid generative counterfactuals which are able to flip the prediction of target model. In the experiments, although we simply consider the DNN as the target prediction model, due to its general good performance for raw data instances, our proposed framework can also be easily applied with other different prediction models. The main contributions of this paper are summarized as follows:
\begin{itemize}[leftmargin=*]
\item We design a general framework to derive counterfactuals for raw data instances by employing generative models, aiming to facilitate the reasoning on model behaviors of black-box DNN; 

\item We develop AIP to iteratively update the parameters of generative models in the attribute-informed latent space, according to the counterfactual loss with regards to the desired prediction label; 
    
\item We evaluate the designed framework with AIP on several real-world datasets including raw texts and images, and demonstrate the superiority both quantitatively and qualitatively. 
\end{itemize}

\section{Preliminaries} 

In this section, we briefly introduce some related contexts to our problem, as well as some basics of the employed techniques. 

\subsection{Counterfactual Explanation} 

Counterfactual explanation is essentially a natural extension under the framework of example-based reasoning~\cite{rissland1991example}, where particular data samples are provided to promote the understandings on model behaviors. Nevertheless, counterfactuals are not common examples for model interpretation, since they are typically generated under the ``what-if'' circumstances which may not necessarily exist. According to the theory proposed by J. Pearl~\cite{pearl2018book}, three distinct levels of cognitive ability are needed to fully master the behaviors of a particular model, i.e., \emph{seeing}, \emph{doing} and \emph{imagining} from the easiest to the hardest. In fact, counterfactual explanation is just raised to meet the imagining-level cognition for model interpretation. 

Within the contexts of this paper, we only discuss counterfactuals under the assumption of ``closest possible world''~\cite{wachter2017counterfactual}, where desired outcomes can be obtained through the smallest changes to the world. To be specific and simple without loss of generality, consider a binary classification model $f_{\boldsymbol\theta}:~\mathbb{R}^{d} \rightarrow \{0,1\}$, where $0$ and $1$ respectively indicate the undesired and desired output. The model input $\mathbf{x} \in \mathbb{R}^{d}$ is further assumed to be sampled from data distribution $\mathcal{P}(\mathbf{x})$. Then, given a query instance $\mathbf{x}_{0}$ with the undesired model output (i.e., $f_{\boldsymbol\theta}(\mathbf{x}_{0})=0$), the corresponding counterfactual $\mathbf{x}^{*}$ can be mathematically represented as:
\begin{equation}\label{eq_cf}
    \mathbf{x}^{*}=\argmin_{\mathbf{x}|\mathcal{P}(\mathbf{x})>\eta} l(\mathbf{x}, \mathbf{x}_{0}) \quad
    \mathrm{s.t.} \ f_{\boldsymbol\theta}(\mathbf{x}^{*})=1,
\end{equation}
where $l: \mathbb{R}^{d} \times \mathbb{R}^{d} \rightarrow \mathbb{R}^{+}$ indicates a distance measure defined in the input space, and $\eta>0$ denotes the threshold which quantifies how likely the sample $\mathbf{x}$ is under the distribution $\mathcal{P}(\mathbf{x})$. The obtained counterfactual $\mathbf{x}^{*}$ is regarded to be valid if it can effectively flip the target classifier $f_{\boldsymbol\theta}$ to the desired prediction. 

Although finding counterfactuals is somewhat similar to generating adversarial examples (in terms that both tasks aim to flip the model decision by minimally perturbing the input instance), they are essentially different in nature. Following the previous settings, the adversarial sample $\mathbf{x}^{adv}$ for model $f_{\boldsymbol\theta}$, with query instance $\mathbf{x}_{0}$, can be generally indicated by:
\begin{equation}\label{eq_adv}
    \mathbf{x}^{adv}=\argmin_{\mathbf{x}=\mathbf{x}_{0}+\boldsymbol\delta} \|\boldsymbol\delta\|_{p} \quad
    \mathrm{s.t.} \ f_{\boldsymbol\theta}(\mathbf{x}^{adv}) \neq f_{\boldsymbol\theta}(\mathbf{x}_{0}),
\end{equation}
where $\boldsymbol\delta$ denotes the adversarial perturbation on the query, $\|\cdot\|$ represents the norm operation and $p \in \{\infty, 1, 2, \cdots\}$. Comparing with Eq.~\ref{eq_cf}, we note that counterfactual example has two significant differences from adversarial sample. First, counterfactual generation process is subject to the original data distribution, while adversarial samples are not constrained by the distribution. This difference brings about the fact that counterfactuals are all in-distribution samples, but adversarial examples are mostly out-of-distribution (OOD) samples. Second, counterfactual changes on the query need to be human-perceptible, while adversarial perturbations are usually inconspicuous~\cite{sen2019should}. Therefore, the key problem of counterfactual explanation actually lies in how to generate such in-distribution sample, with human-perceptible changes on the query, to flip the model decision as desired.

\subsection{Generative Modeling} 

Generative modeling is a typical task under the paradigm of unsupervised learning. Different from discriminative ones, which involves discriminating input samples across classes, generative modeling aims to summarize the data distribution of input variables and further create new samples that plausibly fit into that distribution~\cite{murphy2012machine}. In practice, a well-trained generative model is capable of generating new examples that are not only reasonable, but also indistinguishable from real examples in the problem domain. Conventional examples of generative modeling include Latent Dirichlet Allocation (LDA) and Gaussian Mixture Model (GMM). 

As emerging families of generative modeling, Generative Adversarial Network (GAN)~\cite{goodfellow2014generative} and Variational Auto-Encoder (VAE)~\cite{kingma2013auto} have been attracting lots of attentions due to their exceptional performance in a myriad of applications, especially for the task on image and text generation~\cite{van2016conditional,hu2017toward}. 
By taking full advantage of their power on raw data with high dimensionality, we are able to better investigate how those data samples were created in the first place, which potentially benefits the generation of certain hypothetical example. To this end, we specifically employ some advanced generative models (i.e., GAN and VAE) to study the counterfactual explanation for black-box DNN on raw data instances, providing effective generative counterfactuals for better model understanding.

\section{Counterfactual Generation} 

In this section, we first introduce the designed generative counterfactual framework for raw data instances. Then, we present how to specifically construct the attribute-informed latent space with generative models. Finally, we show the details of our proposed AIP method on how to effectively obtain such counterfactuals.

\subsection{Generative Counterfactual Framework}

\begin{figure}[t] 
\centering
\includegraphics[width=\columnwidth]{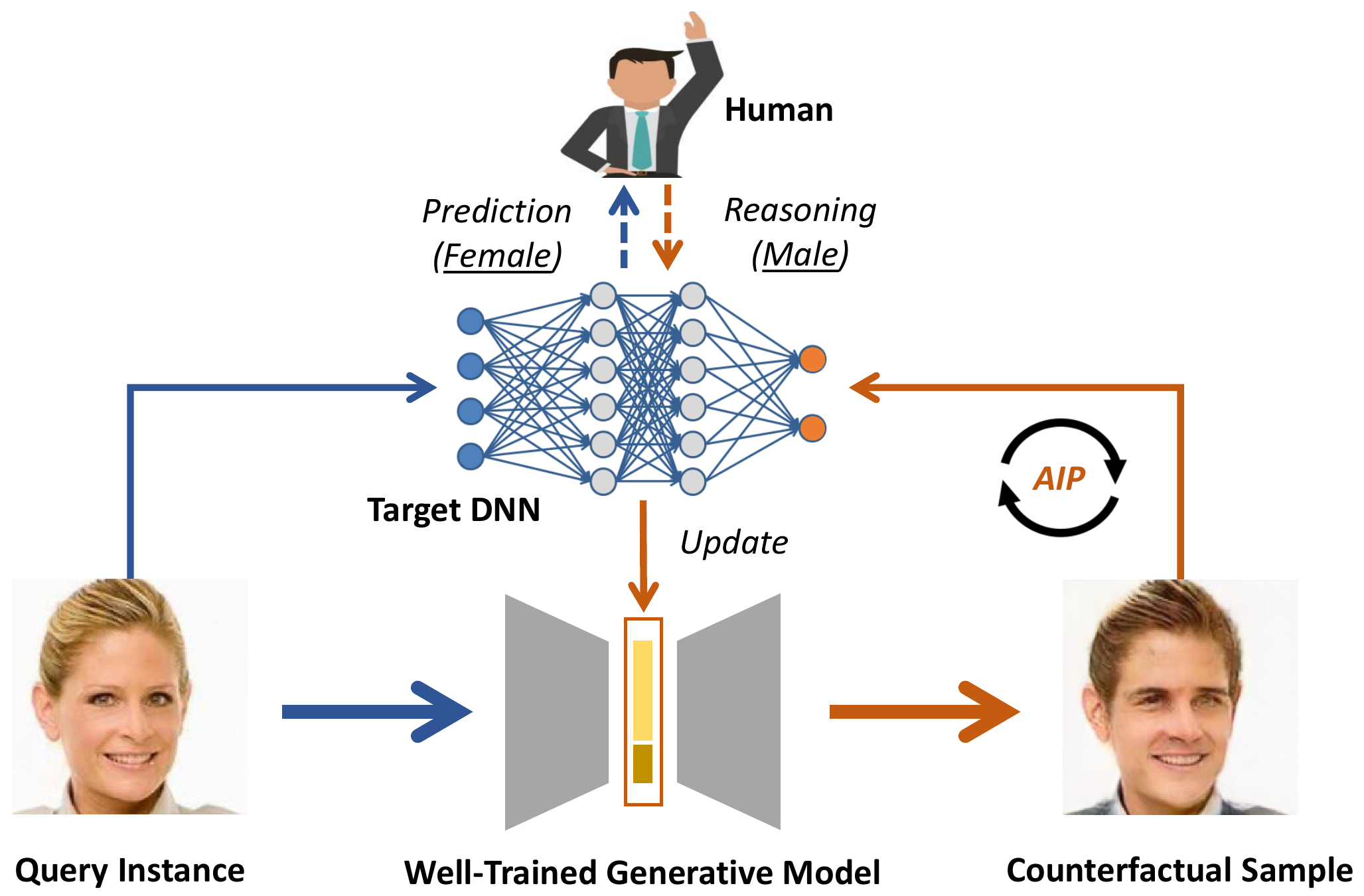}
\caption{Designed framework for counterfactual sample.} 
\label{fig:framework}
\end{figure} 

We design a framework to create counterfactual samples for raw data instances, as illustrated by Fig.~\ref{fig:framework}. To effectively handle the high dimensionality and unsemantic features, we utilize the generative modeling techniques to aid the counterfactual generation process. Consider a target DNN $F_{\boldsymbol\phi}: \mathbb{R}^{d} \rightarrow \{1,\cdots,C\}$, which is the black-box model for counterfactual analysis, where $\mathbb{R}^{d}$ is the input data space and $\{1,\cdots,C\}$ denotes the model prediction space with $C$ different outputs. Given a query instance $\mathbf{x}_{0}$, $F_{\boldsymbol\phi}(\mathbf{x}_{0})=\mathbf{y}_{0}$ outputs a one-hot vector. To effectively generate a valid counterfactual sample $\mathbf{x}^{*}\in\mathbb{R}^{d}$ that can flip the $F_{\boldsymbol\phi}$ decision to $\mathbf{y}^{*}\in\{1,\cdots,C\}$ as desired, a generative model is trained to achieve this in the framework. 
The applied generative modeling actually plays two important roles in the counterfactual generation process: (1) generative modeling guarantees that all created instances are in-distribution samples, since it can be regarded as a stochastic procedure that generates samples $\mathbf{x} \in \mathbb{R}^{d}$ under the particular data distribution $\mathcal{P}(\mathbf{x})$; (2) generative modeling generally assumes that underlying latent variables can be mapped to the data space under certain circumstances, which ensures the sufficient feasibility for hypothetical examples. Thus, a well-trained generative model is the basis for high-quality counterfactuals within the designed framework. 

The employed generative model specifically serves two sub-tasks for counterfactual generation, i.e., data \emph{encoding} and \emph{decoding}. For raw data instances like images, the input space $\mathbb{R}^{d}$ could be extremely large, which makes it difficult and inefficient to directly create counterfactuals for the query. In our designed framework, data encoding is conducted to map the input data space to a low-dimension attribute-informed latent space, which is formulated as a joint embedding space for both raw features and data attributes. In this way, each data sample $\mathbf{x}$ can be effectively encoded through the function $G^{enc}_{\boldsymbol\psi}: \mathbb{R}^{d} \rightarrow \mathbb{R}^{k}\oplus\mathbb{R}^{t}$, where $\mathbb{R}^{k}$ is the latent space for raw feature embeddings, $\mathbb{R}^{t}$ indicates the data attribute space, and $\oplus$ represents a concatenation operator. Reversely, the mapping for decoding is from the attribute-informed latent space to the original data space. The decoder function can be similarly indicated by $G^{dec}_{\boldsymbol\omega}: \mathbb{R}^{k}\times\mathbb{R}^{t} \rightarrow \mathbb{R}^{d}$. Although $G^{enc}_{\boldsymbol\psi}$ and $G^{dec}_{\boldsymbol\omega}$ typically have two different focuses, they are jointly trained as a whole generative model in an end-to-end manner. The issues about how to derive $G^{enc}_{\boldsymbol\psi}$ and $G^{dec}_{\boldsymbol\omega}$ will be particularly discussed in Sec.~\ref{sec_als}. 

To finally obtain the counterfactual sample $\mathbf{x}^{*}$ for model $F_{\boldsymbol\phi}$ with query $\mathbf{x}_{0}$, we further need to modify the attribute-informed latent space of the deployed generative model. Specifically, we use the proposed AIP method to update the attribute-informed latent vector of $\mathbf{x}_{0}$, according to the counterfactual loss calculated. Assuming $G^{enc}_{\boldsymbol\psi}(\mathbf{x}_{0})=\mathbf{z}_{0}\oplus\mathbf{a}_{0}$ ($\mathbf{z}_{0} \in \mathbb{R}^{k}, \mathbf{a}_{0} \in \mathbb{R}^{t}$), AIP method can jointly update $\mathbf{z}_{0}$ and $\mathbf{a}_{0}$, so as to minimize the corresponding loss counter-factually. The overall counterfactual loss consists of two parts, i.e., \emph{prediction loss} and \emph{perturbation loss}. Prediction loss is set to ensure the flip of model decisions, and perturbation loss is involved to guarantee the ``closest possible'' changes on the query, which are both indispensable for counterfactual generation. For the prediction loss, we simply follow the common cross-entropy term, expressed as $L_{d}(F_{\boldsymbol\phi}(\mathbf{x}), \mathbf{y}^{*})=-\mathbf{y}^{*}\log(F_{\boldsymbol\phi}(\mathbf{x}))-(1-\mathbf{y}^{*})\log(1-F_{\boldsymbol\phi}(\mathbf{x}))$. For the perturbation loss, we employ two $l_{2}$ norms respectively on $\mathbb{R}^{k}$ and $\mathbb{R}^{t}$, indicated by $L_{b}(\mathbf{z},\mathbf{a},\mathbf{z}_{0},\mathbf{a}_{0})=\|\mathbf{z}-\mathbf{z}_{0}\|_{2}+\|\mathbf{a}-\mathbf{a}_{0}\|_{2}$ ($\mathbf{z} \in \mathbb{R}^{k}, \mathbf{a} \in \mathbb{R}^{t}$), to restrain the query changes, which can also be regarded as a regularization term as well. Further, the overall counterfactual loss can thus be represented as follows:
\begin{equation}
    L_{c}(\mathbf{z}, \mathbf{a}, \mathbf{z}_{0}, \mathbf{a}_{0}, \mathbf{y}^{*})= L_{d}\left(F_{\boldsymbol\phi}(G^{dec}_{\boldsymbol\omega}(\mathbf{z}, \mathbf{a})), \mathbf{y}^{*}\right)+\alpha L_{b}(\mathbf{z},\mathbf{a},\mathbf{z}_{0},\mathbf{a}_{0}),
\label{eq_cfl}
\end{equation}
where $\alpha$ is a balance coefficient between the two loss terms. With the proposed AIP method, the designed framework can generate the valid counterfactual example $\mathbf{x}^{*}$ with the aid of optimized $\mathbf{z}^{*}, \mathbf{a}^{*}$ through the decoder function (i.e., $\mathbf{x}^{*}=G^{dec}_{\boldsymbol\omega}(\mathbf{z}^{*}, \mathbf{a}^{*})$). The details of the proposed AIP method will be introduced in Sec.~\ref{sec_attrpp}.

\subsection{Attribute-Informed Latent Space}
\label{sec_als}

\begin{figure}[t] 
\centering
\includegraphics[width=0.9\columnwidth]{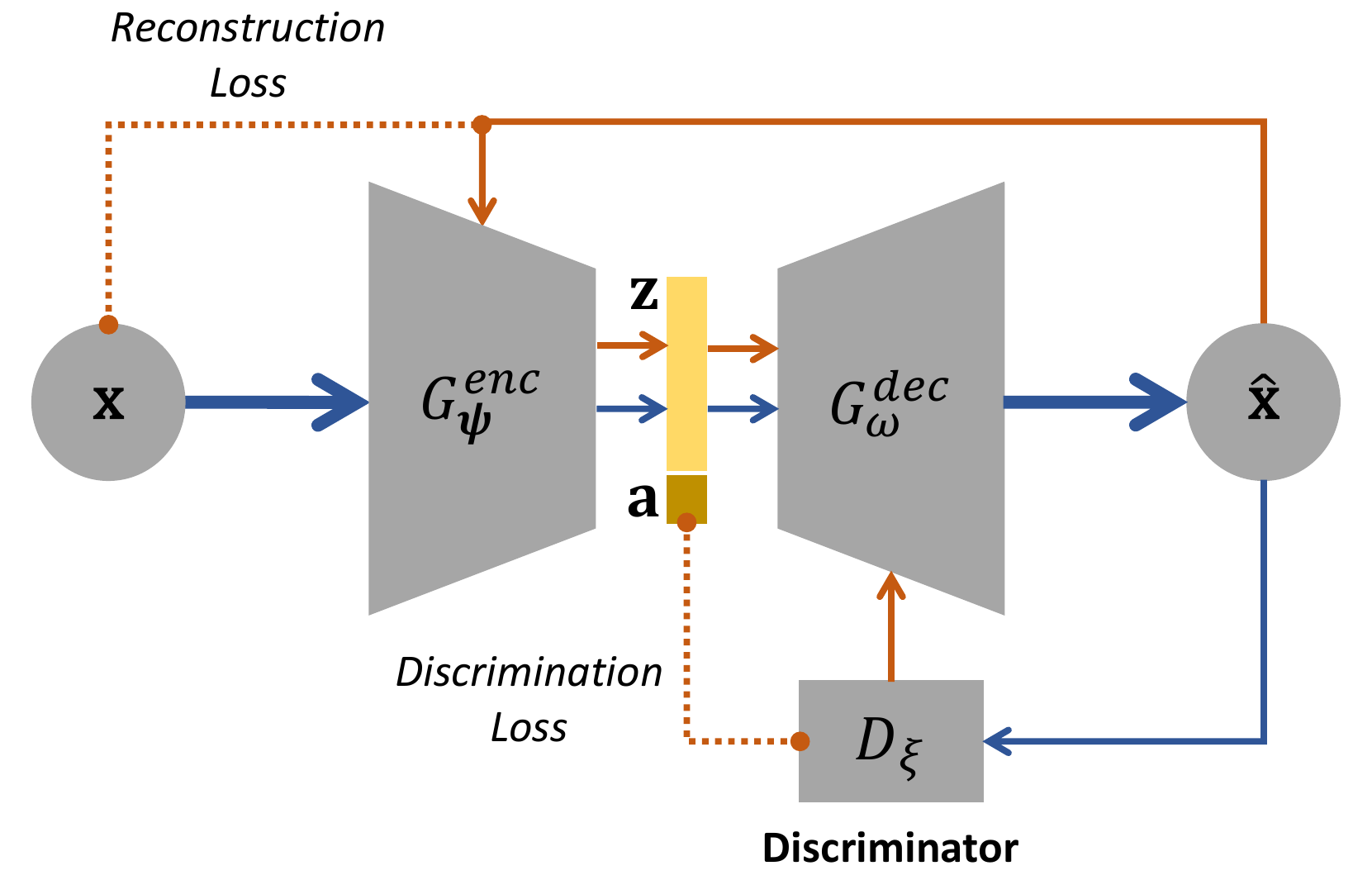}
\caption{General illustration of the attribute-informed latent space in generative models. Particularly, blue arrows indicate the forward flow of computations, while orange arrows indicate the back-propagation flow of gradients. The dash lines denote the losses for generative model training.} 
\label{fig:space}
\end{figure} 

Constructing an appropriate attribute-informed latent space is the key part for generative modeling in our designed framework, which has direct influences on the quality of generated counterfactuals. To achieve this, we need to well train a generative model, better capturing the raw data features as well as relevant data attributes, where embedded features can bring about more robust bases for counterfactual analysis, and incorporated attributes are able to provide more semantics for conditional generation. Here, the data attributes mainly indicate those extra information from humans along with raw instances, such as annotations or labels, which can usually be represented as one-hot vectors. 

In practice, it is common that different generative models are employed for different tasks or data. Since different models typically involve disparate architectures, their training schemes can totally differ from each other. Take the GAN and VAE for example, where GAN is usually trained to obtain an equilibrium between a generator and a discriminator function, while VAE is typically trained to maximize a variational lower bound of the data log-likelihood. Therefore, to better introduce how to specifically construct the attribute-informed latent space with generative models, we present a general illustration shown by Fig.~\ref{fig:space}, although it may not be fully representative for all kinds of models. 

We generally introduce the modeling process with an encoder-decoder structure, which corresponds to the data encoding and decoding in our designed framework. Essentially, the attribute-informed latent space can be regarded as an extended code space of auto-encoders. By concatenating attribute vector $\mathbf{a}$ to raw feature embedding $\mathbf{z}$, the decoder function aims to achieve the conditional generation based on $\mathbf{a}$. To ensure the attribute consistency between original sample $\mathbf{x}$ and generated sample $\hat{\mathbf{x}}$, discriminator $D_{\xi}$ is particularly employed, which is trained separately and used to classify the attributes of $\hat{\mathbf{x}}$. To effectively train such generative model, two basic loss terms are required, which are the reconstruction loss and discrimination loss. The overall training can be indicated by:
\begin{multline}
    \min_{\psi,\omega} \quad  \underbrace{\underset{\substack{\mathbf{x}\sim \mathcal{P}(\mathbf{x}) \\ \mathbf{a}\sim \mathcal{P}(\mathbf{a})}}{\mathbb{E}} \ \sum_{i=1}^{t} -a_{i}\log\left(D_{\xi}^{i}(\hat{\mathbf{x}})\right)-(1-a_{i})\log\left(1-D_{\xi}^{i}(\hat{\mathbf{x}})\right)}_{\text{Discrimination Loss}}  \\ + \underbrace{\underset{\mathbf{x}\sim \mathcal{P}(\mathbf{x})}{\mathbb{E}} \|\mathbf{x}-\hat{\mathbf{x}}\|_{2}}_{\text{Reconstruction Loss}},
\end{multline} 
where $a_{i}$ denotes the $i$-th attribute in $\mathbf{a}$, and $D_{\xi}^{i}$ indicates the prediction of $D_{\xi}$ on the $i$-th attribute. After sufficient training, $G^{enc}_{\boldsymbol\psi}$ and $G^{dec}_{\boldsymbol\omega}$ can be effectively obtained, and the attribute-informed latent space can be further constructed with the aid of $G^{enc}_{\boldsymbol\psi}$. For specific tasks and architectures, the generative modeling process could be further enhanced with more specific losses or other advanced tricks.

\subsection{Attribute-Informed Perturbation}
\label{sec_attrpp}

With the obtained $G^{enc}_{\boldsymbol\psi}$ and $G^{dec}_{\boldsymbol\omega}$ for generative modeling, we then introduce the proposed AIP method to finally derive the counterfactual for target DNN $F_{\boldsymbol\phi}$ with the query $\mathbf{x}_{0}$. To guarantee the quality of the generated counterfactuals, AIP needs to find the sample that can minimize the counterfactual loss indicated by Eq.~\ref{eq_cfl}. Under the ``closest possible world'' assumption, the corresponding counterfactual sample can be denoted as:
\begin{equation}
    \mathbf{x}^{*}=G^{dec}_{\boldsymbol\omega}\left(\argmin_{\mathbf{z}\in \mathbb{R}^{k}, \ \mathbf{a}\in \mathbb{R}^{t}} L_{c}(\mathbf{z}, \mathbf{a}, \mathbf{z}_{0}, \mathbf{a}_{0}, \mathbf{y}^{*})\right).
\label{eq_cfs}
\end{equation}
To effectively solve Eq.~\ref{eq_cfs}, the proposed AIP method utilizes an iterative gradient-based optimization algorithm with dynamic step sizes (controlled by a decaying factor $\beta$), which helps the iteration process converge faster. In each iteration, the updated $\mathbf{z}$ and $\mathbf{a}$ can be derived as follows:
\begin{equation}
\left\{
\begin{aligned}
\mathbf{z}^{(n+1)} = \mathbf{z}^{(n)}-\mu^{(n)}\nabla_{\mathbf{z}} L_{c}\left(\mathbf{z}^{(n)}, \mathbf{a}^{(n)}, \ \mathbf{z}_{0}, \ \mathbf{a}_{0}, \ \mathbf{y}^{*}\right) \\
\mathbf{a}^{(n+1)} = \mathbf{a}^{(n)}-\gamma^{(n)}\nabla_{\mathbf{a}} L_{c}\left(\mathbf{z}^{(n)}, \mathbf{a}^{(n)}, \ \mathbf{z}_{0}, \ \mathbf{a}_{0}, \ \mathbf{y}^{*}\right)
\end{aligned}
\right.
,
\label{eq_update}
\end{equation} 
where $n$ indicates the iteration index, $\mu$ and $\gamma$ respectively denotes the step sizes of updates on $\mathbf{z}$ and $\mathbf{a}$. Specifically, the proposed AIP method can be summarized in Algorithm~\ref{algo}. It is important to note that AIP only works on the optimization of $\mathbf{z}, \mathbf{a}$, and does not involve the parameter update on $F_{\boldsymbol\phi}$, $G^{enc}_{\boldsymbol\psi}$, $G^{dec}_{\boldsymbol\omega}$. Thus, the proposed AIP method should be less time-consuming and easily deployed for counterfactual generation task, compared with those generative frameworks which need extra model training~\cite{samangouei2018explaingan,singla2019explanation}.  

\begin{algorithm}[t]
\KwInput{$F_{\boldsymbol\phi}$, $G^{enc}_{\boldsymbol\psi}$, $G^{dec}_{\boldsymbol\omega}$, $\mathbf{x}_{0}$, $\mathbf{y}^{*}$, $\mu$, $\gamma$, $\alpha$, $\beta$, $n_{\max}$}

\KwOutput{Counterfactual sample $\mathbf{x}^{*}$}

 Initialize $\mu$, $\gamma$, $\alpha$, $\beta$ \;
 
 Initialize $n=0, \ \mathbf{x}=\mathbf{x}_{0}$ \;
 
 Construct the latent space with $\mathbf{z}\oplus\mathbf{a} \leftarrow G^{enc}_{\boldsymbol\psi}(\mathbf{x})$ \;
 
 \While{$F_{\boldsymbol\phi}(G^{dec}_{\boldsymbol\omega}(\mathbf{z}, \mathbf{a}))\neq \mathbf{y^{*}} \ \mathrm{or} \ n \leq n_{\max}$}{

 Update $\mathbf{z}$ and $\mathbf{a}$ according to Eq.~\ref{eq_update} \;
 
 Update step sizes with $\mu \leftarrow \beta\mu$ and $\gamma \leftarrow \beta\gamma$ \;
 
 $n \leftarrow n+1$
 }
 
 Reconstruct the optimized sample with $\mathbf{x}^{*} \leftarrow G^{dec}_{\boldsymbol\omega}(\mathbf{z}, \mathbf{a})$ \;
 
 \eIf{$F_{\boldsymbol\phi}(\mathbf{x}^{*})==\mathbf{y}^{*}$}{
 
 \textbf{Return} $\mathbf{x}^{*}$ as the counterfactual for $F_{\boldsymbol\phi}$ with query $\mathbf{x}_{0}$;
 
 }{
 \textbf{Return} None -- No valid counterfactual exists;
 }
 
\caption{Attribute-Informed Perturbation (AIP)}
\label{algo}
\end{algorithm}

\section{Experiments} 

In this section, we evaluate the designed counterfactual generation framework with the proposed AIP method on several real-world datasets, both quantitatively and qualitatively. Overall, we mainly conduct two sets of experiments respectively on \emph{text} and \emph{image} counterfactual generation, by utilizing different data modeling techniques. With all conducted experiments, we aim to answer the following four key research questions:
\begin{itemize}[leftmargin=*]
\item How \emph{effective} is the designed framework in generating counterfactuals with AIP, regarding to different types of raw data?

\item How is the \emph{quality} of created counterfactuals from our designed framework aided by AIP, comparing with other methods?

\item How \emph{efficient} is the counterfactual generation under the designed framework with AIP, comparing with other potential ways? 

\item Can we further \emph{benefit} other practical tasks with the counterfactuals generated from our design framework with AIP method? 
\end{itemize}

\subsection{Experimental Settings} 

\subsubsection{Real-World Datasets.} 

Throughout the whole experiments, we employ three real-world datasets to evaluate the performance of the designed framework with AIP method, including both raw texts and images. The relevant data attributes depend on the particular tasks, which are collected either from labels or annotations. The statistics of the involved datasets are shown in Table~\ref{tab:data}. 

\begin{table}[t] 
\caption{Dataset statistics in experiments.}
\vspace{-0.3cm}
\centering
\setlength{\tabcolsep}{3.9pt} 
\begin{normalsize}
\begin{tabular}[width=1cm]{cccccc}
\toprule
Datasets        & \#Instance      & \#Attribute     & Type         &  Domain \\
\midrule
Yelp             & $455,000$         & 1          & Raw Texts      & Sentiment   \\  
Amazon             & $558,000$       & 1          & Raw Texts      & Sentiment         \\  
CelebA            & $202,599$        & 13         & Raw Images           & Face Feature   \\
\bottomrule
\end{tabular}
\end{normalsize}
\label{tab:data}
\end{table}

\begin{itemize}[leftmargin=*]
\item {\textbf{Yelp User Review Dataset}}\footnote{\url{https://www.yelp.com/dataset}}~\cite{asghar2016yelp}: This dataset consists of user reviews from the Yelp associated with relevant rating scores. We involve a tailored and modified version of this data for our experiments on text counterfactual generation. Specifically, we consider the reviews with ratings higher than three as \emph{positive} samples and regard the others as \emph{negative} ones, and we further use these sentiment labels as the relevant attribute for data modeling. The vocabulary of our involved Yelp data contains more than $9,000$ words, and the average review length is around $9$ words. 

\item {\textbf{Amazon Product Review Dataset}}\footnote{\url{http://jmcauley.ucsd.edu/data/amazon/index_2014.html}}~\cite{he2016ups}: This dataset is also involved as a raw textual dataset for our experiments. Similar to the Yelp data, we modify the original rating information of reviews into the sentiment categories (i.e., \emph{positive} and \emph{negative}), and further model these labels as an sentiment attribute of the raw textual reviews. Amazon dataset has more than $50,000$ words in vocabulary, and the average length is around $15$. 

\item {\textbf{CelebFaces Attributes (CelebA) Dataset}}\footnote{\url{http://mmlab.ie.cuhk.edu.hk/projects/CelebA.html}}~\cite{liu2015deep}: This is a large-scale face attributes dataset, containing tons of raw face images with human annotations. We employ this dataset for our experiments on image counterfactual generation, and select out $13$ representative face attributes (out of 40) for data modeling along with raw face images. The involved thirteen attributes include: \emph{Male}, \emph{Young}, \emph{Blond\_Hair}, \emph{Pale\_Skin}, \emph{Bangs}, \emph{Mustache} and etc.   
\end{itemize}

\subsubsection{Target Model for Interpretation.} 

Since we mainly discuss the counterfactuals of raw data instances, DNN is a better choice as our target model. For the target DNN $F_{\boldsymbol\phi}$ in our experiments, we employ some regular structures for the corresponding tasks. Particularly, for the text sentiment classification, we use a common convolutional architecture in~\cite{kim2014convolutional} to pre-train a DNN classifier for further counterfactual analysis. For the image attribute classification task, similarly, we utilize a simple convolutional network~\cite{guo2017simple} to prepare a target classifier, where the model is trained with one of those attributes as the label. During the evaluations on counterfactual generation, target DNNs are fixed without further training.

\subsubsection{Employed Generative Modeling Techniques.} 

In our experiments, different data modeling techniques are employed for different types of data. In particular, we use different generative models to construct the corresponding attribute-informed latent space, regarding to text and image data. For textual reviews (Yelp, Amazon), we utilize the modeling techniques introduced in~\cite{hu2017toward}, and build a transformer-based VAE to effectively formulate the relevant attribute-informed latent space. For face images (CelebA), we mainly follow the modeling method of AttGAN~\cite{he2019attgan}, where more complicated training schemes are employed, compared with the general one shown in Sec.~\ref{sec_als}, for better visual quality of the generated images. Both of the employed generative models should be well-trained on the corresponding datasets before the counterfactual generation process, so as to guarantee the high quality of generated counterfactuals.

\subsubsection{Alternative Methods and Baselines.} 

To effectively evaluate the performance of the designed framework with AIP, we incorporate following alternative methods and baselines for comparison. 

\begin{itemize}[leftmargin=*]
\item {\textbf{TextBugger}~\cite{li2018textbugger}}: This is a general method for adversarial text generation, which is built based on the word attribution and bug selection. The created text samples can effectively flip the prediction of the target classifier. We employ this method as a baseline to specifically compare with our generated text counterfactuals. 

\item {\textbf{DeepWordBug}~\cite{gao2018black}}: This is another method focusing on the adversarial text generation, where a token scoring strategy is utilized to guide the character-level adversarial perturbation. This method is employed as a baseline for text counterfactuals as well. 

\item {\textbf{FGSM}~\cite{goodfellow2015explaining}}: Fast gradient sign method is a common way to generate image adversarial samples, by using the gradients of the loss with respect to the input. The sample created by this method can effectively maximize the loss, so as to flip the original prediction. We employ this method as a baseline specifically for our generated image counterfactuals. 

\item {\textbf{Counter\_Vis}~\cite{goyal2019counterfactual}}: This is a recent method in generating image counterfactuals, where particular image regions are replaced to flip the model decision. We employ this method as an alternative method for image counterfactual generation. 

\item {\textbf{CADEX}~\cite{moore2019explaining}}: This is a state-of-the-art method for counterfactual generation, where the gradient-based method is directly applied to modify the input space of query. This method is originally proposed for tabular data, and we modify it simply as an alternative for image counterfactuals, due to the particularity of texts.

\item {\textbf{xGEMs}~\cite{joshi2018xgems}}: This is a state-of-the-art method for generating counterfactuals, which also employs the generative modeling technique for sample generation. This method only involves the latent space modeling and cannot achieve the conditional generation with semantic attributes. We employ this method as an important alternative for both text and image counterfactuals. 

\item {\textbf{AIP\_R}}: This is the random version of our proposed AIP method, which updates all parameters in a random way. 
\end{itemize}

\subsection{Implementations} 

In this part, we introduce the implementation details for our experiments on text and image counterfactual generation, corresponding to the following Sec.~\ref{text_exp} and Sec.~\ref{image_exp}. 

\subsubsection{Details for Text Counterfactuals} 

\paragraph{\textbf{Hyper-parameters.}}
We implement all related algorithms and models by the PyTorch framework\footnote{https://pytorch.org/}. Specifically for text counterfactuals, we set the balance coefficient $\alpha=0.8$ in Eq.~\ref{eq_cfl} during the main evaluations. The decaying factor is set as $\beta=0.95$, and the maximum iteration is set as $n_{\max}=300$. As for the initial step sizes for optimization, we set $\mu=1$ and $\gamma=2$ in the related experiments.

\paragraph{\textbf{Target DNN model $F_{\mathbf{\phi}}$ for texts.}} 
We train two CNN classifiers respectively on Yelp and Amazon datasets as our target models, following the architectures introduced in work~\cite{kim2014convolutional}. In particular, both CNN employ the \emph{CNN-non-static} version of training, as illustrated in the paper. The deployed target CNN for Yelp data has $89.58\%$ testing accuracy, and the deployed CNN for Amazon has $88.16\%$ testing accuracy.

\paragraph{\textbf{Generative model $G^{enc}_{\psi},G^{dec}_{\omega}$ for texts.}}
We employ a transformer based VAE to conduct the relevant data modeling. Specifically, the overall structure of the employed \emph{Encoder} and \emph{Decoder} can be illustrated by the following Tab.~\ref{tab:tave}. The size of transformer and latent space are both set as $256$. As for the training phase, we use another classifier, trained separately with $F_{\mathbf{\phi}}$, as the discriminator $D_{\xi}$. The relevant batch size is $128$, embedding dropout rate is $0.5$, and learning rate is $10^{-3}$. 

\begin{table}[h]
\renewcommand\arraystretch{1.1}
\centering
\caption{Structure for the employed transformer-based VAE.}
\vspace{-0.3cm}
\label{tab:tave}
\begin{tabular}{|c||c|}
\hline
\textbf{Encoder} (\emph{from top to down}) & \textbf{Decoder}  (\emph{from top to down}) \\
\hline
Embedding Layer & Multi-Head Attention Layers \\
Multi-Head Attention Layers & Addition \& Normalization \\
Addition \& Normalization & Dense Layer \\
Dense Layer & Addition \& Normalization \\
Addition \& Normalization & Multi-Head Attention Layers \\
Multi-Head Attention Layers & Fully-Connected Layer \\
GRU Layer & Softmax \\
Summation & /  \\
\hline
\end{tabular}
\end{table}

\subsubsection{Details for Image Counterfactuals} 

\paragraph{\textbf{Hyper-parameters.}}
Similar as text counterfactuals, we also use PyTorch to implement relevant models and algorithms. In particular, we set $\alpha=1.5$ in Eq.~\ref{eq_cfl}. The decaying factor and maximum iteration in Algorithm~\ref{algo} is respectively set as $\beta=0.9$ and $n_{\max}=500$. Besides, the initial optimization step sizes is set as $\mu=2$ and $\gamma=3$. 

\paragraph{\textbf{Target DNN model $F_{\mathbf{\phi}}$ for images.}} 
We train a basic CNN on the CelebA dataset as the target according to the architecture in~\cite{guo2017simple}, specifically using the ``Male'' annotation as the label for training. Essentially, the target CNN here is a binary gender classifier which predicts an input image as either ``Male'' or ``Female''. The deployed CNN has the testing accuracy with $95.67\%$. 

\paragraph{\textbf{Generative model $G^{enc}_{\psi},G^{dec}_{\omega}$ for images.}}
The generative modeling for CelebA data largely follows the training schemes in~\cite{he2019attgan}. Our modeling attributes include: "Pale\_Skin", "Bangs", "Black\_Hair", "Blond\_Hair", "No\_Beard", "Brown\_Hair", "Bushy\_Eyebrows", "Male", "Eyeglasses", "Young", "Mustache", "Bald",  "Mouth\_Slightly\_Open". Particularly, we present our employed structure of \emph{Encoder} and \emph{Decoder} in the following Tab.~\ref{tab:attgan}. Here, DeConvolution indicates the transposed operation on convolution, and (64,4,2) respectively denotes the dimension, kernel size and stride, for example. From the structure, we note that the latent size is $1,024$. For the training phase, we specifically set the batch size as $32$, and learning rate as $2\times10^{-4}$. The discriminator $D_{\xi}$ is a multi-class classifier, trained separately from $F_{\mathbf{\phi}}$.

\begin{table}[h]
\renewcommand\arraystretch{1.1}
\centering
\caption{Structure for the employed AttGAN.}
\vspace{-0.3cm}
\label{tab:attgan}
\begin{tabular}{|c||c|}
\hline
\textbf{Encoder} (\emph{from top to down}) & \textbf{Decoder} (\emph{from top to down}) \\
\hline
Convolution Layer (64,4,2) & DeConvolution Layer (1024,4,2) \\
Normalization & Normalization \\
Convolution Layer (128,4,2) & DeConvolution Layer(512,4,2) \\
Normalization & Normalization \\
Convolution Layer (256,4,2) & DeConvolution Layer(256,4,2) \\
Normalization & Normalization \\
Convolution Layer (512,4,2) & DeConvolution Layer(128,4,2) \\
Normalization & Normalization \\
Convolution Layer (1024,4,2) & DeConvolution Layer(3,4,2) \\
Normalization &  /  \\
\hline
\end{tabular}
\end{table}

\subsection{Text Counterfactual Evaluations}\label{text_exp}

In this part, we evaluate the experimental results of the designed framework with AIP in generating text counterfactuals, regrading to a convolutional neural network (CNN) built for sentiment classification. The involved raw texts for target DNN come from the user/product reviews in Yelp and Amazon datasets, where $90\%$ are used for training, $5\%$ for development and $5\%$ for testing. 

\subsubsection{Effectiveness Evaluation.}\label{text_effect}

\begin{figure}[t] 
\centering
\includegraphics[width=0.9\columnwidth, height=2.5cm]{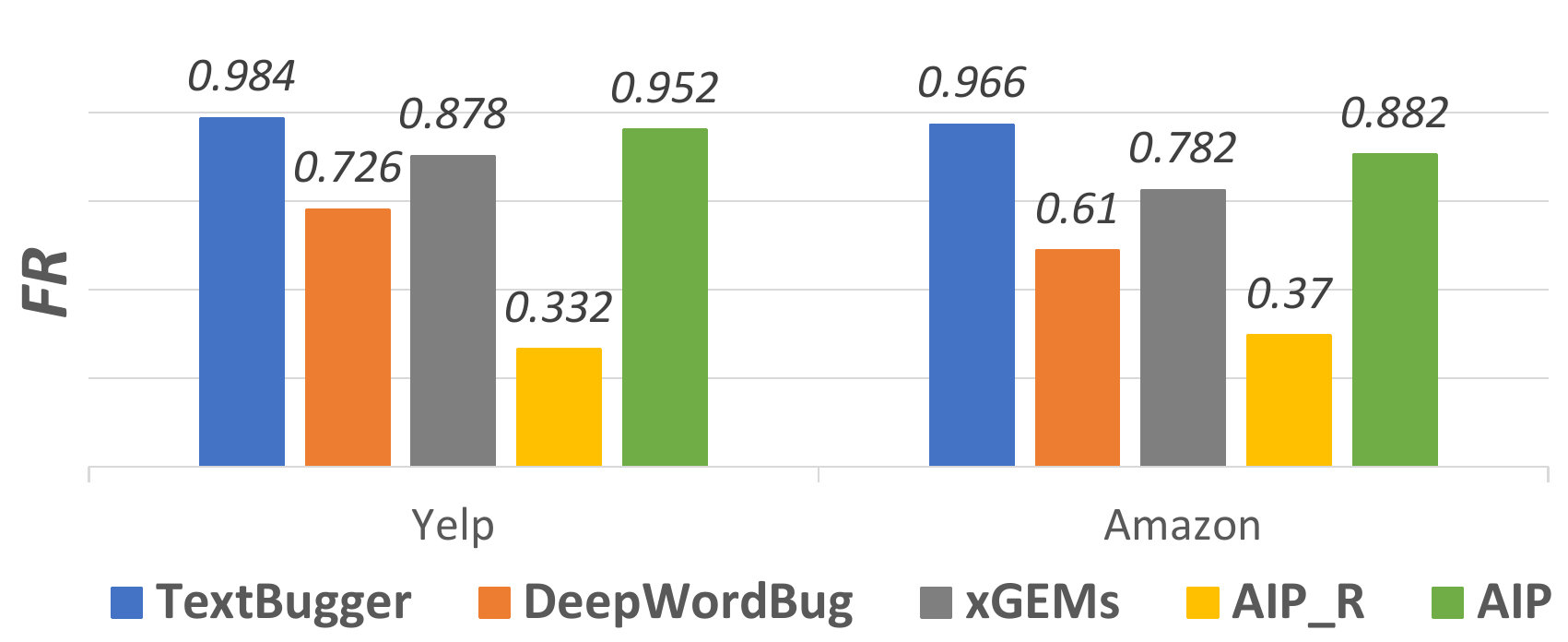}
\vspace{-0.2cm}
\caption{Effectiveness evaluation for text counterfactuals.} 
\label{fig:text_effect}
\end{figure} 

In order to evaluate the effectiveness for text counterfactuals, we employ the metric \emph{Flipping Ratio} (FR) to measure the relevant performance, which reflects how likely the generated text samples would flip the model decision to $\mathbf{y}^{*}$. Specifically, FR can be calculated as follows:
\begin{equation}
    \mathrm{FR}=|\mathcal{X}_{f}| \Big/ |\mathcal{X}_{q}| \quad \left(\mathbf{x}_{0}\in \mathcal{X}_{f} \ \mathrm{if} \ F_{\boldsymbol\phi}(\mathbf{x}_{0})=\mathbf{y}^{*} \right),
\label{eq_fr}
\end{equation}
where $\mathcal{X}_{f}$ indicates the set of query samples with which new flipping instances can be generated by particular methods, and $\mathcal{X}_{q}$ denotes the set of all testing queries. In our experiments, there are $500$ testing queries in total (i.e., $|\mathcal{X}_{q}|=500$), which are randomly selected from the test set. Fig.~\ref{fig:text_effect} illustrates our experimental results on both Yelp and Amazon datasets. According to the numerical results, we note that our designed framework with AIP can work well on both datasets, and has competitive performance among all other alternatives as well as baselines, although TextBugger achieves the highest FR score with better robustness. Besides, we also observe that AIP\_R does not effectively work for generating flipping samples, which indicates that random optimization in attribute-informed latent space cannot help for counterfactual sample generation.

\subsubsection{Quality Evaluation.}\label{text_qual}

\begin{figure}[t] 
\centering
\includegraphics[width=0.9\columnwidth, height=2.5cm]{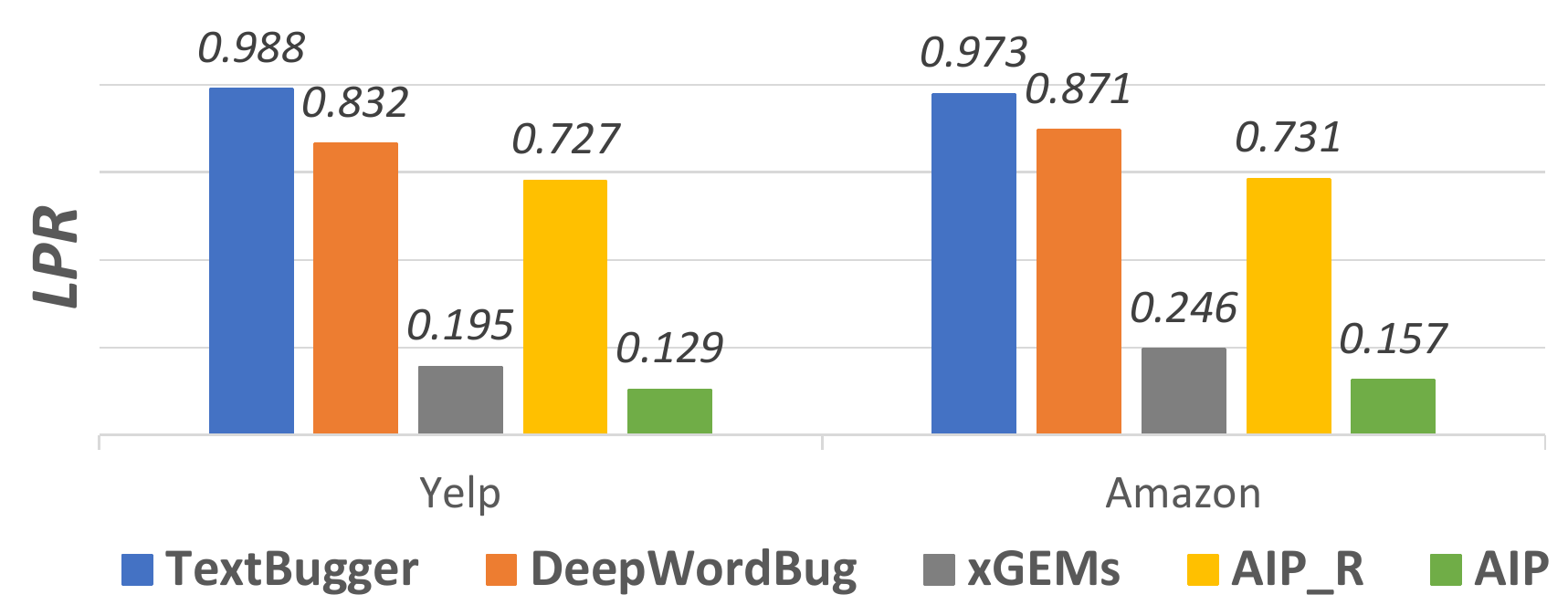}

\caption{Quality evaluation for text counterfactuals.} 
\label{fig:text_qual}
\end{figure} 

As for the quality assessment of counterfactual samples, we employ the \emph{Latent Perturbation Ratio} (LPR) metric to measure the latent closeness between the generated sample $\mathbf{x}^{*}$  and original query instance $\mathbf{x}_{0}$. Since high-quality counterfactual samples typically need to ensure sparse changes in the robust feature space, thus the smaller the LPR is, the better the counterfactual we have. To be specific, the LPR can be calculated by:   
\begin{equation}
    \mathrm{LPR}=\big\|\mathbf{z}^{*}-\mathbf{z}_{0}\big\|_{0} \Big/k,
\label{eq_lpr}
\end{equation}
where $\|\cdot\|_{0}$ indicates the $l_{0}$ norm operation, $\mathbf{z}^{*}$ and $\mathbf{z}_{0}$ are the raw feature embeddings respectively for $\mathbf{x}^{*}$ and $\mathbf{x}_{0}$. To make a fair comparison, we use the same encoder function $G^{enc}_{\boldsymbol\psi}$ for all generated samples to obtain the corresponding latent representation vectors. In this set of experiments, the latent dimension is $256$ (i.e., $k=256$), and the final LPR value for particular method is recorded with the average over $500$ testing queries. The relevant numerical results are presented in Fig.~\ref{fig:text_qual}. From the experiments, it is noted that xGEMs and the proposed AIP method significantly outperform other baselines, indicating that the corresponding generated samples actually maintain more robust features regarding to the query. Furthermore, the proposed AIP is noted to be slightly better than xGEMs, which may partially result from the conditional generation brought by attribute vector $\mathbf{a}$. This set of results also validate a fact that adversarial samples typically utilize some artifacts to flip the model decisions, instead of using some robust features.

\subsubsection{Efficiency Evaluation.}\label{text_effi}

\begin{figure}[t] 
\centering
\includegraphics[width=0.9\columnwidth, height=2.5cm]{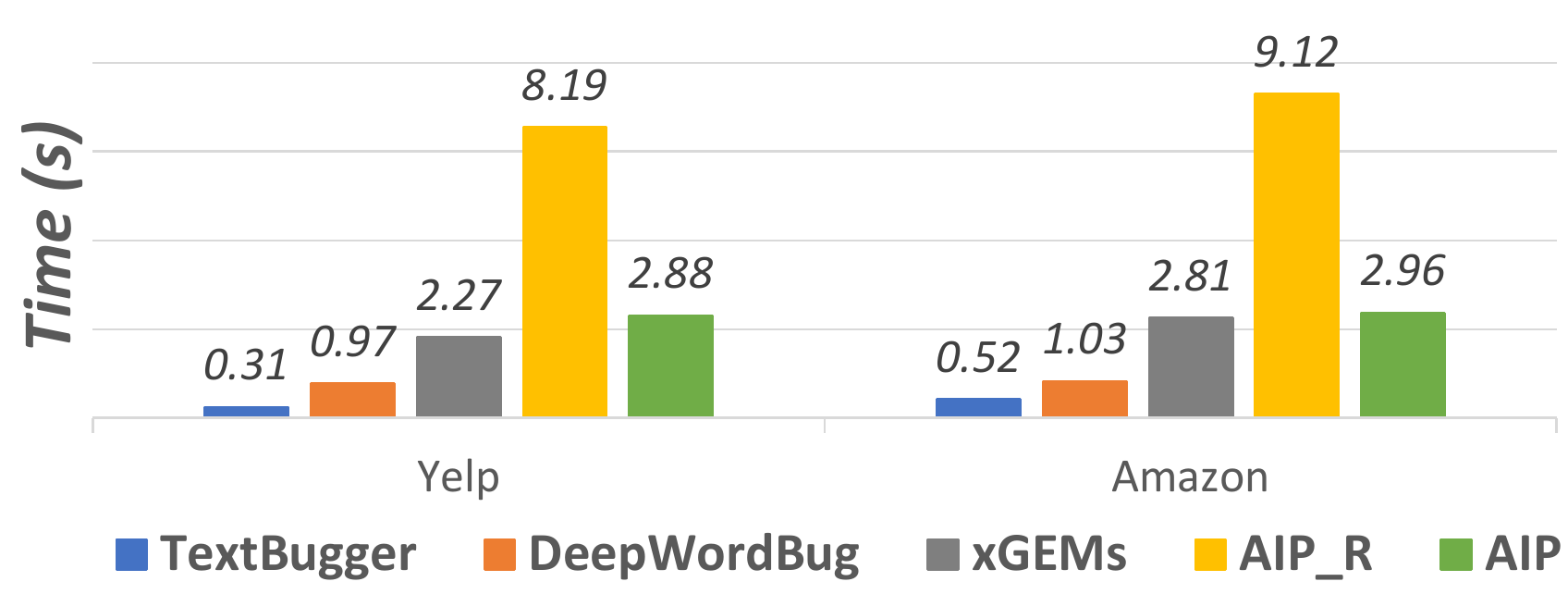}

\caption{Efficiency evaluation for text counterfactuals.} 
\label{fig:text_effi}
\end{figure} 

To compare the efficiency, we record the time consumption for each method over $500$ testing queries in the generation phase on the same machine. Specifically, we calculate the average time cost for one query, and further employ this as the metric to access the efficiency for particular methods. Fig.~\ref{fig:text_effi} shows the relevant experimental results. Based on the statistics, it is observed that adversarial related methods (i.e., TextBugger and DeepWordBug) typically consume less time per query in average, compared with the counterfactual generation methods, which is mainly due to the fact that adversarial methods do not need to conduct encoding computations before sample generation. As for our proposed AIP method, the time efficiency is roughly the same as the alternative xGEMs, but it is significantly better than its random version AIP\_R which needs more iterations to converge.

\subsubsection{Qualitative Case Studies.}\label{text_case}

\begin{table}[]
\centering
\caption{Case studies on generated text samples.}
\vspace{-0.3cm}
\label{tab:textcf} 
\resizebox{\columnwidth}{!}{%
\begin{tabular}{l}
\hline
Counterfactual on \emph{Negative} sentiment (Yelp) \\ 
\textbf{Query}: \texttt{this is the worst walmart neighborhood market out of any of them} \\ \hline
\textbf{TextBugger}: \texttt{this is the worst \textcolor{blue}{wa1mart neighborho0d} market out of \textcolor{blue}{a ny} of them} \\ 
\textbf{DeepWordBug}: \texttt{this \textcolor{blue}{id} the \textcolor{blue}{wosrt} walmart \textcolor{blue}{neighobrhood} market out of any of \textcolor{blue}{htem}} \\ 
\textbf{xGEMs}: \texttt{\textcolor{blue}{that} is \textcolor{blue}{good} walmart market out of any \textcolor{blue}{neighborhood}} \\ 
\textbf{AIP}: \texttt{this is the \textcolor{blue}{best} walmart neighborhood market \textcolor{blue}{for all} of them} \\ 
\hline 

\\

\hline

Counterfactual on \emph{Positive} sentiment (Amazon) \\ 
\textbf{Query}: \texttt{this item works just as i thought it would} \\ \hline
\textbf{TextBugger}: \texttt{this item \textcolor{blue}{w0rks} just as i \textcolor{blue}{tho ught} it \textcolor{blue}{wou1d}} \\ 
\textbf{DeepWordBug}: \texttt{this item \textcolor{blue}{wroks} just \textcolor{blue}{ae} i thought it \textcolor{blue}{wolud}} \\ 
\textbf{xGEMs}: \texttt{this item works \textcolor{blue}{out poorly} just as i thought \textcolor{blue}{disappointed}} \\ 
\textbf{AIP}: \texttt{this item works \textcolor{blue}{bad} just as i thought it would \textcolor{blue}{not play}} \\ \hline
\end{tabular}
}
\end{table}

\begin{figure*}[t] 
\centering
\includegraphics[width=0.9\textwidth, height=2.9cm]{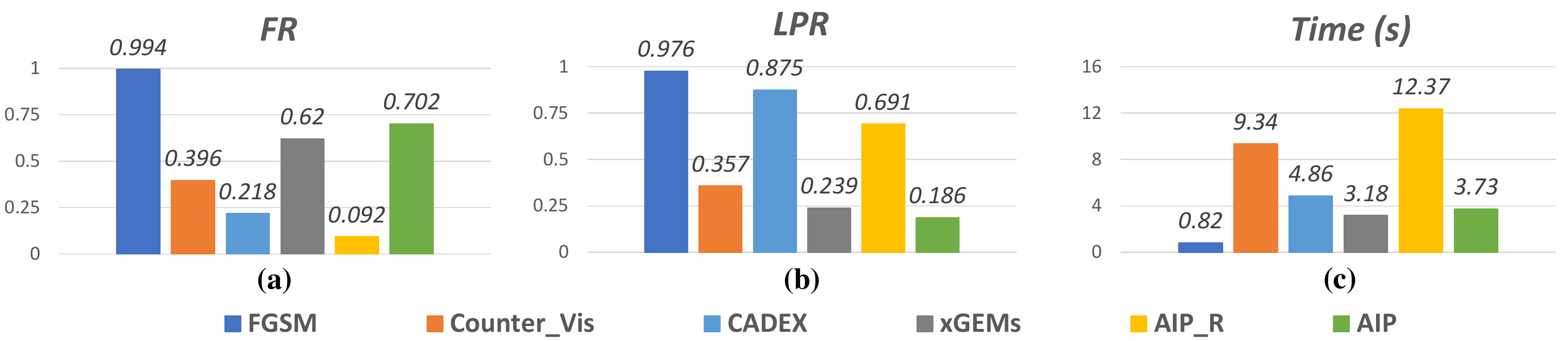}
\vspace{-0.2cm}
\caption{Evaluations for image counterfactual generation.}
\label{fig:imageall}
\end{figure*} 

\begin{figure*}[t] 
\centering
\includegraphics[width=\textwidth]{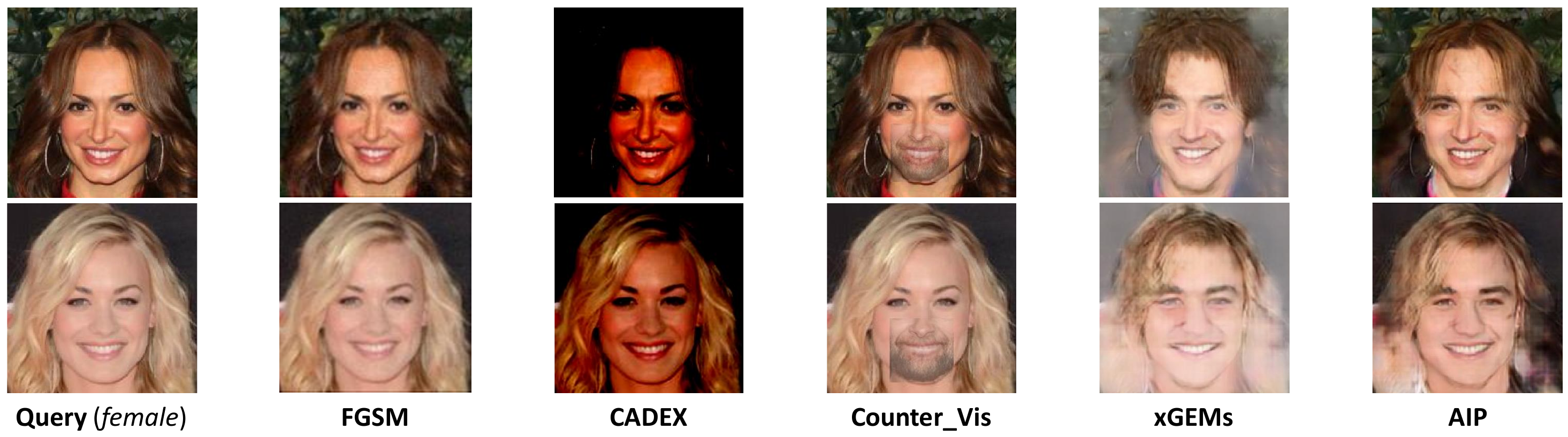}
\vspace{-0.5cm}
\caption{Qualitative case studies on generated image samples.}
\label{fig:imagecase}
\end{figure*}

Here, we present several representative case studies from different methods, shown in Tab.~\ref{tab:textcf}, aiming to provide a qualitative comparison for generated text samples. Based on the Tab.~\ref{tab:textcf}, we can see that adversarial texts typically provide limited insights for humans on counterfactual analysis, since they mainly make use of the model artifacts to flip the prediction. Nevertheless, with the samples generated by xGEMs and AIP, we can easily observe some sentiment variation regarding to the query instance, which sheds light on model behaviors and facilitates further human reasoning on black-box models. Besides, compared with xGEMs, the proposed AIP method usually can generate more sensible counterfactuals with the aid of attribute conditions.

\subsection{Image Counterfactual Evaluations}\label{image_exp}

In this part, we specifically evaluate the designed framework with AIP on image counterfactual generation. Instead of simply considering one attribute for conditional generation in texts, we take multiple attributes into account for image counterfactuals. In this set of experiments, our target DNN follows the common CNN architecture and is trained as a gender classifier, which can classify an input image as \emph{Male} or \emph{Female}. All involved raw images for target DNN come from the CelebA dataset, and we use $90\%$ data for training, $5\%$ for development, $5\%$ for testing. The relevant quantitative results are all illustrated by Fig.~\ref{fig:imageall}.

\subsubsection{Effectiveness Evaluation.}\label{image_effe}

For the effectiveness assessment, we still use the FR metric indicated by Eq.~\ref{eq_fr}. In the experiments, we set $|\mathcal{X}_{q}|=500$, and aim to test how many of them can be effectively flipped with particular methods. Fig.~\ref{fig:imageall}(a) illustrates the relevant numerical results, where adversarial method FGSM performs the best on FR and can flip nearly every testing query. We note that the proposed AIP method ranks the second, and outperforms other counterfactual generation methods. Besides, it is also observed that CADEX and AIP\_R performs relatively bad for the image counterfactual task within certain iterations, even though CADEX is proved to work well for tabular instances~\cite{moore2019explaining}.

\subsubsection{Quality Evaluation.}\label{image_qual}

Similar to text counterfactual scenario, we employ the LPR metric, shown as Eq.~\ref{eq_lpr}, to measure the quality of the generated image counterfactuals. In experiments, the latent dimension $k$ constructed by $G^{enc}_{\boldsymbol\psi}$ is $1,024$ (i.e., $k=1024$), and the corresponding LPR for particular method is recorded by calculating the average over $500$ testing queries. Relevant experimental results are shown by Fig.~\ref{fig:imageall}(b). Based on the LPR comparison, we note that the samples generated by FGSM and CADEX change a lot in the latent feature space, because both methods directly rely on the input perturbation for sample generation. As for the proposed AIP, it achieves the lowest LPR among all the alternatives and baselines, and it is significantly better than its random version AIP\_R.

\subsubsection{Efficiency Evaluation.}\label{image_effi}

We similarly employ the average time consumption per query to evaluate the efficiency aspect for image counterfactual generation. Specifically, the average time is obtained over the $500$ testing queries randomly selected from the test set. Fig.~\ref{fig:imageall}(c) shows the relevant experimental results. According to the statistics and comparison, we note that FGSM is the most efficient one, and xGEMs consumes the least time in average among all other counterfactual-based methods. As for the proposed AIP, a competitive efficiency performance is observed, and is remarkably superior compared with that of Counter\_Vis, CADEX and AIP\_R.

\subsubsection{Qualitative Case Studies.}\label{image_case}

To facilitate a qualitative comparison among different methods, we specifically show some case studies, illustrated by Fig.~\ref{fig:imagecase}. We select several query instances whose model predictions are female, and then employ different methods to generate the corresponding image samples which flip the model decisions for counterfactual purpose. According to the results, we note that the samples generated by FGSM and CADEX do not have salient visual changes regarding to the query instances, which largely limits the human reasoning on model behaviors. Among other alternative methods, it is observed that the proposed AIP is capable of generating counterfactuals with better visual quality, which present much smoother transitions from female to male.

\subsection{Influence of Hyper-parameter $\alpha$}
\begin{figure}[h] 
\centering
\includegraphics[width=\columnwidth, height=3cm]{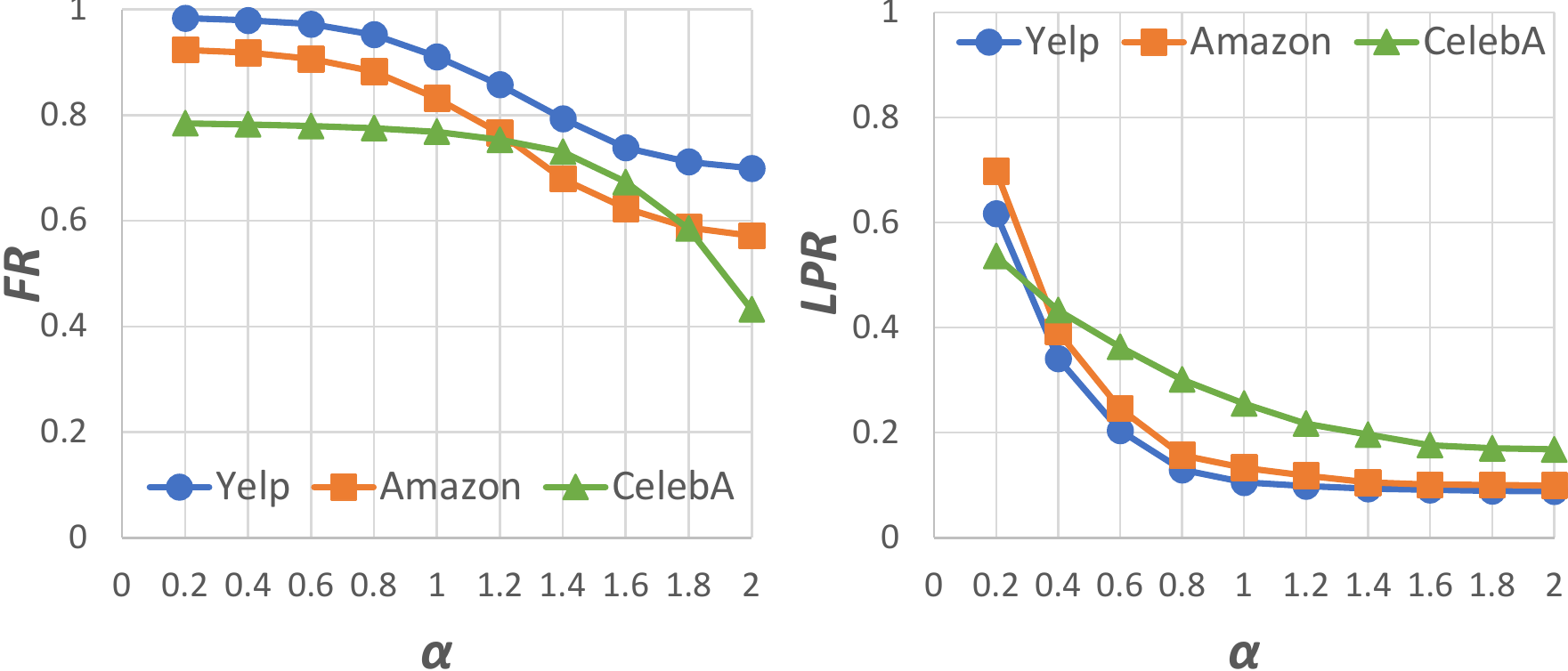}
\vspace{-0.5cm}
\caption{Influence of $\alpha$ on FR and LPR metrics.} 
\label{fig:app}
\end{figure}

In this part, we show some additional results on the influence of hyper-parameter $\alpha$ in Eq.~\ref{eq_cfl}. Other experimental settings keep unchanged. The relevant results are shown by Fig.~\ref{fig:app}. 

Based on the results, we observe that $\alpha$ serves as a knob to control the effectiveness and sample quality of the designed framework. To select an appropriate $\alpha$, we actually need to strike a balance between FR and LPR, where the larger the $\alpha$ is, the lower the effectiveness is and the higher the sample quality is. Different data types may also have different trade-off curves.

\subsection{Applications} 

In this part, we focus on some practical scenarios which may benefit from the counterfactual samples generated by our designed framework. In particular, we show the applications of the framework respectively on \emph{feature interaction} and \emph{data augmentation}.

\subsubsection{Feature Interaction}

\begin{figure}[t] 
\centering
\includegraphics[width=\columnwidth]{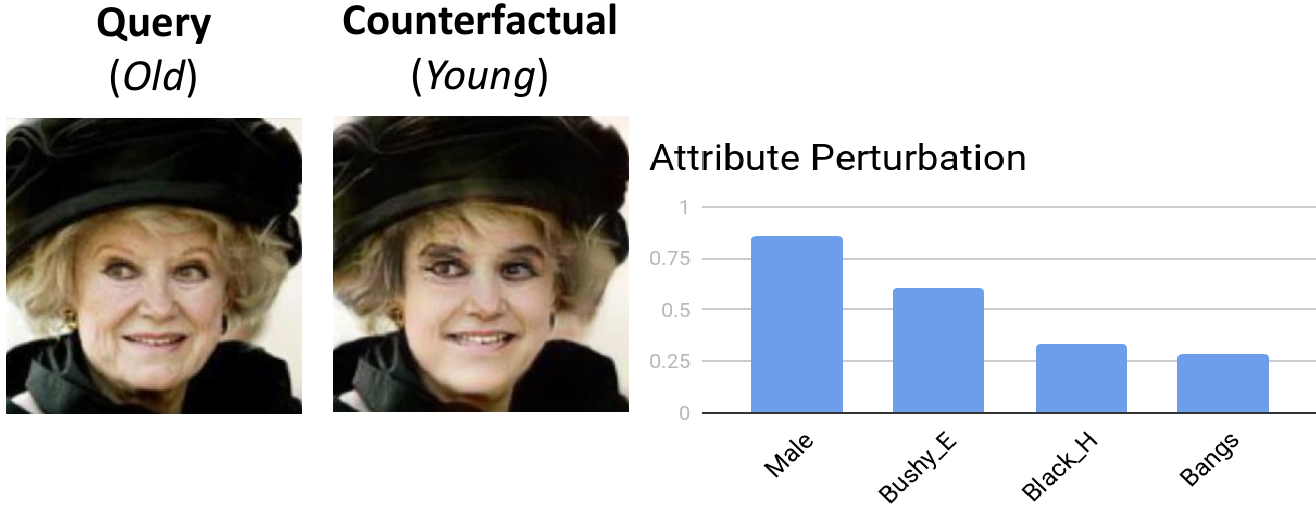}
\vspace{-0.5cm}
\caption{Feature interactions for the decision change.} 
\label{fig:task1}
\end{figure} 

Understanding the feature interaction could be very important in lots of real-world domains. A typical example is the bias detection task, where humans aim to find out a related set of features which can significantly influence the correctness or fairness of model decision. Utilizing our designed framework for counterfactual analysis can partially help this practical task. By observing the perturbation scale on attribute vector $\mathbf{a}$ of the generated counterfactual, humans can have a sense on which semantic features contribute significantly to the flipping of model decision. To illustrate the point, we show another case result from the designed framework with AIP in Fig.~\ref{fig:task1}. Here, we train an age classifier on the CelebA dataset as our target DNN, and aim to analyze the feature interaction of a query prediction as ``Old''. Based on the attribute perturbations of the generated sample, we note that the top semantic attributes are ``Male'', ``Bushy\_Eyebrows'', ``Black\_Hair'' and ``Bangs'', besides the target attribute. This result directly demonstrates the fact that the ``Male'' attribute has a strong interaction with the predicted attribute for this particular query, and the target DNN exists potential gender bias for its age predictions.

\subsubsection{Data Augmentation} 

\begin{table}[t]
\centering
\caption{Model performance with data augmentation.}
\vspace{-0.3cm}
\label{tab:task2}
\begin{tabular}{|cc|c|c|}
\hline
\multicolumn{2}{|c|}{\textbf{Dataset}} & \textbf{CNN}~\cite{kim2014convolutional} & \textbf{VDCNN}~\cite{conneau2017very} \\
\hline
\multirow{2}{*}{\textbf{Yelp}} & \textit{Initial} & 82.33\% ($\pm$ 0.61\%) & 88.79\% ($\pm$ 0.53\%) \\
 & \textit{Augmented} & 83.16\% ($\pm$ 0.57\%) & 89.95\% ($\pm$ 0.46\%) \\
\hline
\multirow{2}{*}{\textbf{Amazon}} & \textit{Initial} & 81.96\% ($\pm$ 0.52\%) & 88.55\% ($\pm$ 0.63\%) \\
 & \textit{Augmented} & 82.41\% ($\pm$ 0.49\%) & 88.76\% ($\pm$ 0.55\%) \\
\hline
\end{tabular}

\begin{tabular}{|cc|c|c|}
\hline
\multicolumn{2}{|c|}{\textbf{Dataset}} & \textbf{CNN}~\cite{guo2017simple} & \textbf{ResNet}~\cite{he2016deep} \\
\hline
\multirow{2}{*}{\textbf{CelebA} \ } & \textit{Initial} & 87.32\% ($\pm$ 0.22\%) & 90.96\% ($\pm$ 0.27\%) \\
 & \textit{Augmented} & 88.85\% ($\pm$ 0.21\%) & 91.35\% ($\pm$ 0.25\%) \\
\hline
\end{tabular}
\end{table} 

Another application of the designed framework is the data augmentation for model training. By taking full advantage of the generated counterfactual samples as new training instances, we aim to obtain better DNN models with higher performance and robustness. Specifically, to test the improvement, we train several DNN models on relatively smaller training sets, which are essentially the subsets of original data. For the sentiment classifiers on Yelp and Amazon, our initial training size is $20,000$, containing $10,000$ positive and $10,000$ negative reviews. The extra counterfactual training size is $2,000$ whose queries are randomly selected from the initial training set. For the binary age classifier on CelebA, we employ a similar setting for training, where each class includes $10,000$ initial samples, and $2,000$ generated counterfactual samples are further incorporated for augmentation. Relevant experimental results are shown in Tab.~\ref{tab:task2}. Based on the statistics, we note that the augmented training with counterfactual samples typically achieves higher classification accuracies with smaller variances, which can also be observed under some advanced DNN structures.

\section{Related Work}

Generating counterfactual explanation is just one of many interpretation methods for black-box models, which generally belongs to the family of interpretable machine learning. According to the particular problems they focus on, interpretation methods can be divided into the following three categories in general. 

The first category of methods aims to answer the ``\emph{What}''-type questions, i.e., what part of the input mostly contribute to the model prediction. A representative work in this category is LIME~\cite{ribeiro2016should}, where authors specifically employ linear models to approximate the local decision boundary and further formulate it as a sub-modular optimization problem for model interpretation. The feature importance in LIME is essentially obtained by observing the prediction changes after perturbing input samples. Similar related methods can also be found in Anchors~\cite{ribeiro2018anchors} and SHAP~\cite{lundberg2017unified}. Another common methodology under this category is to utilize the model gradient information, where gradients are typically regarded as an indicator for perturbation sensitivity. Related methods can be found in GradCAM~\cite{selvaraju2017grad}, Integrated Gradients~\cite{sundararajan2017axiomatic}, and SmoothGrad~\cite{smilkov2017smoothgrad}. 

The second category aims to answer the ``\emph{Why}''-type questions, i.e., why the input is predicted as label \emph{A} instead of \emph{B}. The methods under this category can be quite different from the previous ones, since these methods basically need to consider two labels simultaneously. There are several different methodologies proposed for this problem. For example, the authors in~\cite{dhurandhar2018explanations} design a contrastive perturbation method to derive related positive and negative features of input regarding to the concerned label. Besides, a general method based on structural causal models is proposed in~\cite{miller2018contrastive} to tackle the problem specifically in classification and planning scenarios. Also, a generative framework CDeepEx is designed in~\cite{feghahati2018cdeepex} to particularly investigate this problem for images by utilizing GAN. 

The third category lies in the ``\emph{How}''-type questions, i.e., how to particularly modify the input so as to flip the model prediction to the preferred label. This problem is a natural extension of the ``Why''-type, and it can somewhat to be handled by the second category of methods under some simple scenarios. However, for problems with high-dimension space, previous categories of methods typically fail due to the intractable computation for sample modification. Several particular methods are raised to solve this issue. For example, authors in~\cite{goyal2019counterfactual} propose a straightforward solution with image region replacement, which is essentially a feature replacement process for input with the aid of a distractor. In work~\cite{agarwal2019removing}, authors novelly use the input itself as the distractor for feature replacement by utilizing GAN for inpainting. Besides, generative modeling is another potential way for this problem, and related methods can be found in~\cite{singla2019explanation,joshi2018xgems,liu2019generative}. Our work belongs to this branch of methodology.

\section{Conclusion And Future Work} 

In this paper, we design a framework to generate counterfactual explanation for black-box DNN models specifically with raw data instances. By taking advantage of the generative modeling techniques, we effectively construct an attribute-informed latent space for particular data, and further utilize this space for counterfactual generation. To guarantee the validity of the generated samples, we propose the AIP method to iteratively optimize the specific attribute-informed latent vectors according to the counterfactual loss term, from which the counterfactuals can be finally obtained through data reconstruction. We evaluate the designed framework with AIP on several real-world datasets, including both texts and images, and demonstrate its effectiveness, sample quality as well as efficiency. Future extension of this work may possibly include the investigation under the ``close possible worlds'' assumption, where the goal is to find an optimal set of counterfactuals for a query instead of a single sample. Besides, employing causal models for counterfactual generation is another promising direction to explore.


\bibliographystyle{ACM-Reference-Format}
\bibliography{paper_ref}

\end{document}